\documentclass{article}

     \usepackage[preprint,nonatbib]{neurips_2020}

\usepackage[utf8]{inputenc} 
\usepackage[T1]{fontenc}    
\usepackage{hyperref}       
\usepackage{url}            
\usepackage{booktabs}       
\usepackage{amsfonts}       
\usepackage{nicefrac}       
\usepackage{microtype}      

\usepackage{algorithm}
\usepackage{algorithmic}
\usepackage{amsmath}
\usepackage{bm}
\usepackage{color}
\usepackage{subcaption}
\usepackage{graphicx}
\usepackage{multirow}
\usepackage{url}

\newcommand{\kETAL}        {\emph{et al.}}
\newcommand{\bx}        {\textbf{x}}
\newcommand{\by}        {\textbf{y}}
\newcommand{\bD}        {\textbf{D}}
\newcommand{\mD}        {\mathcal{D}}
\newcommand{\btheta}    {\bm{\theta}}
\newcommand{\mT}        {\mathcal{T}}
\newcommand{\mL}        {\mathcal{L}}

\title{Adversarial Meta-Learning}

\author{
Chengxiang Yin,\textsuperscript{\rm 1}
Jian Tang,\textsuperscript{\rm 2}
Zhiyuan Xu,\textsuperscript{\rm 1}
Yanzhi Wang,\textsuperscript{\rm 3} \\
\textsuperscript{\rm 1}Syracuse University,
\textsuperscript{\rm 2}DiDi AI Labs,
\textsuperscript{\rm 3}Northeastern University\\
cyin02@syr.edu, tangjian@didiglobal.com, zxu105@syr.edu \\
yanz.wang@northeastern.edu \\
}

\begin{document}

\maketitle

\begin{abstract}
Meta-learning enables a model to learn from very limited data to undertake a new task.
In this paper, we study the general meta-learning with adversarial samples.
We present a meta-learning algorithm, ADML (ADversarial Meta-Learner), which leverages clean and adversarial samples to optimize the initialization of a learning model in an adversarial manner.
ADML leads to the following desirable properties:
1) it turns out to be very effective even in the cases
with only clean samples;
%
%
2) it is robust to adversarial samples, i.e., unlike other meta-learning algorithms, it only leads to a minor performance degradation when there are adversarial samples;
3) it sheds light on tackling the cases with limited and even contaminated samples.
%
%
It has been shown by extensive experimental results that ADML consistently outperforms
three representative meta-learning algorithms
in the cases involving adversarial samples,
on two widely-used image datasets, MiniImageNet and CIFAR100,
in terms of both accuracy and robustness.
\end{abstract} 
\section{Introduction}
\label{Sec:Intro}
Deep learning has made tremendous successes and emerged as a \emph{de facto} approach
in many application domains,
such as computer vision and natural language processing,
which, however, depends heavily on huge amounts of labeled training data.
The goal of meta-learning is to enable a model (especially a Deep Neural Network (DNN))
to learn from only a small number of data samples to undertake a new task,
which is critically important to machine intelligence but turns out to be very challenging.
Currently, a common approach to learn is to train a model to undertake a task from scratch
without making use of any previous experience.
Specifically, a model is initialized randomly and then updated slowly using gradient descent with a large number of training samples.
This kind of time-consuming and data-hungry training process is quite different from the way how a human learns quickly from only a few samples and obviously cannot meet the requirement of meta-learning.
Several methods~\cite{finn2017model,vinyals2016matching,snell2017prototypical,sung2018learning} have been proposed to address meta-learning by fixing the above issue.
For example, a well-known work~\cite{finn2017model} presents a novel meta-learning algorithm called MAML (Model-Agnostic Meta-Learning),
which trains a model’s initial parameters carefully such that it has the maximal performance on a new task after its parameters are updated through one or just a few gradient steps with a small amount of new data.
This method is claimed to be model-agnostic since it can be directly applied to any learning model that can be trained with a gradient descent procedure.

Robustness is another major concern for machine intelligence.
It has been shown by~\cite{szegedy2014intriguing} that learning models can be easily fooled by adversarial manipulation of actual training data to cause incorrect classifications.
Therefore, adversarial samples pose a serious security threat to learning tasks,
which need to be properly and effectively handled by learning models and training algorithms.
We show via experiments that existing meta-leaning algorithms (such as MAML~\cite{finn2017model}, Matching Networks~\cite{vinyals2016matching} and Relation Networks~\cite{sung2018learning}) are also vulnerable to adversarial samples,
%
i.e., adversarial samples can lead to a significant performance degradation
for meta-learning.
An adversarial approach, called MetaGAN, was presented in~\cite{zhang2018metagan} for few-shot classification, nevertheless,
in addition to learning better decision boundaries by leveraging fake data with the power of GAN~\cite{goodfellow2014generative},
it paid no attention to deal with the cases involving adversarial samples.
To the best of our knowledge, none of existing works on meta-learning have
well addressed adversarial samples, which, however, are the main focus of this paper.

In this paper, we extend meta-learning to a whole new dimension by studying how to quickly train a model (especially a DNN) for a new task using a small dataset with both clean and adversarial samples.
Since both meta-learning and adversarial learning have been studied recently,
a straightforward solution is to simply combine an existing meta-learning algorithms (e.g., MAML~\cite{finn2017model}) with adversarial training (e.g.,~\cite{Goodfellow2015}).
However, we show such a approach does not work well by our experimental results.
%
We present a novel ADversarial Meta-Learner (ADML), which utilizes antagonistic correlations between clean and adversarial samples to let the inner gradient update arm-wrestle with the meta-update
to obtain a good and robust initialization of model parameters.
Hence, ``adversarial'' in ADML refers to not only adversarial samples but also the way of updating the learning model.
The design of ADML leads to several desirable properties.
First, it turns out to be very effective even in the cases with only clean samples.
%
%
Second, unlike other meta-learning algorithms,
ADML is robust to adversarial samples since it only suffers from a minor performance degradation when encountering adversarial samples,
and it consistently outperforms three representative meta-learning algorithms~\cite{finn2017model,vinyals2016matching,sung2018learning} in such cases.
Most importantly, it opens up an interesting research direction and sheds light on dealing with the cases with limited and even contaminated samples, which are common in real life.
We conducted a comprehensive empirical study for performance evaluation using
two widely-used image datasets,
MiniImageNet~\cite{vinyals2016matching} and CIFAR100~\cite{krizhevsky2009learning}.
Experimental results well justify the effectiveness and superiority of
ADML in terms of both accuracy and robustness.
%

%
%
%
\section{Related Work}
\label{Sec:Related}
%
%
\textbf{Meta-Learning:}
Research on meta-learning has a long history, which can be traced back to some early works~\cite{naik1992meta,Thrun1998}.
%
Meta-learning, a standard methodology to tackle few-shot learning problems,
%
has recently attracted extensive attention due to its important roles in achieving human-level intelligence.
Several specialized
models~\cite{vinyals2016matching,koch2015siamese,snell2017prototypical,sung2018learning} have been proposed for meta-learning,
particularly for 
few-shot classification, by comparing similarity among data samples.
Specifically,
Koch~\kETAL~\cite{koch2015siamese} leveraged a Siamese Networks to rank similarity between input samples and predict if two samples belong to the same class.
In addition, Relation Networks~\cite{sung2018learning} was proposed to classify query images by computing relation scores, which can be extended to few-shot learning.
In~\cite{vinyals2016matching}, Vinyals~\kETAL~presented a neural network model, Matching Networks, which learn
an embedding function and use the cosine distance in an attention kernel to measure similarity.
Another work~\cite{snell2017prototypical} leveraged a similar approach to few-shot classification but used the Euclidean distance
with their embedding function.

Another popular approach to meta-learning is to develop a meta-learner to optimize key
hyper-parameters (e.g., initialization) of the learning model.
Specifically, Finn~\kETAL~\cite{finn2017model} presented a model-agnostic meta-learner, MAML,
to optimize the initialization of a learning model with the objective of maximizing its performance on a new task after updating its parameters with a small number of samples.
Several other methods~\cite{andrychowicz2016learning,Ravi2017,santoro2016meta,Mishra2018} utilize an additional neural network, such as LSTM, to serve as the meta-learner.
A seminal work~\cite{andrychowicz2016learning} developed a meta-learner based on LSTMs and showed how the design of an optimization algorithm
can be cast as a learning problem.
%
%
Ravi~\kETAL~\cite{Ravi2017} proposed another LSTM-based meta-learner to learn a proper parameter update and a general initialization for the learning model.
%
%
Compared to LSTM, a neural network~\cite{santoro2016meta}
is equipped with a large external memory (such as Neural Turing Machine (NTM)~\cite{graves2014neural}),
%
%
which has also been leveraged for meta-learning.
A recent work~\cite{Mishra2018} presented a class of simple and generic meta-learners that use a novel combination of temporal convolutions and soft attention.
%

\textbf{Adversarial Learning:}
DNN models have been shown to be vulnerable to adversarial samples.
Particularly, Szegedy~\kETAL~\cite{szegedy2014intriguing} showed that they can cause a DNN to misclassify an image by applying a certain hardly perceptible perturbation, and moreover, the same perturbation can cause a different network
(trained on a different subset of the dataset)
to misclassify the same input.
It has also been shown by Goodfellow~\kETAL~in~\cite{Goodfellow2015} that injecting adversarial samples during training can increase the robustness of DNN models.
In~\cite{rozsa2016accuracy}, Rozsa~\kETAL~conducted experiments on various adversarial sample generation methods with multiple deep Convolutional Neural Networks (CNNs), and found that adversarial samples are mostly transferable across similar network topologies, and better learning models are less vulnerable.
%
%
The authors of
~\cite{Papernot2017} introduced the first practical demonstration
of a black-box attack controlling a remotely hosted DNN without either the model internals or its training data.
%
%
More recently, Kurakin~\kETAL~\cite{Kurakin2017} studied adversarial learning at scale by proposing an algorithm to train a large scale model,
Inception v3, on the ImageNet dataset,
which has been shown to significantly increase the robustness against adversarial samples.
In addition, the authors of~\cite{Miyato2017} extended adversarial training to the text domain by applying perturbations to word embeddings in an RNN rather than to the original input itself, which has been shown to achieve the state-of-the-art results on multiple benchmark semi-supervised and purely supervised tasks.
%

To the best of our knowledge, meta-learning has not been studied in the setting with
adversarial samples. We not only show a straightforward solution does not work well
but also present a novel and effective method, ADML.

\section{Adversarial Meta-Learning}
\label{Sec:ADML}

\subsection{Problem Statement}
\label{Sec:Problem}
The regular machine learning problem seeks a model that maps
observations $\bx$ to output $\by$; and a training algorithm optimizes
the parameters of the model with a training dataset, whose generalization
is then evaluated on a testing dataset.
While in the setting of meta-learning, the learning model is expected to be trained
with limited data to be able to adapt to a new task quickly.
%
%
Meta-learning includes meta-training and meta-testing.
In the \emph{meta-training}, we use a set $\mT$ of $T$ tasks, each of which
has a loss function $\mL_{i}$, and a dataset $\mD_i$ (with limited data) that is further
split into $\bD_{i}$ and $\bD'_{i}$ for training and testing respectively.
For example, in our experiments, each task is a 5-way classification task.

We aim to develop a meta-learner (i.e., a learning algorithm) that takes as input
the datasets $\mD=\{\mD_1, \cdots, \mD_T\}$ and returns a model with parameters $\btheta$
that maximizes the average classification accuracy on the corresponding
testing sets $\mD'=\{\bD'_1, \cdots, \bD'_T\}$. Note that here these testing data are also
used for meta-training.
In the \emph{meta-testing}, we evaluate the generalization of the learned model with parameters $\btheta$
on new tasks, whose corresponding training and testing datasets may include adversarial samples.
The learned model is expected to learn quickly from just one (1-shot) or $K$ ($K$-shot) training samples
of a new task and deliver highly-accurate results on its testing samples.
An ideal meta-learner is supposed to return a learning model that can deal with new tasks with
only clean samples; and suffers from only a minor performance degradation for new tasks with
adversarial samples.
Note that we only consider classification here since so far only
classification has been addressed in the context of adversarial learning~\cite{Kurakin2017}.
We believe the proposed ADML can be easily extended to other scenarios as long as adversarial
samples can be properly generated.
%
%
%
%
%
%
%

\subsection{Adversarial Meta-Learner (ADML)}
\label{Sec:Algorithm}
We formally present the proposed ADML as Algorithm~\ref{Alg:ADML} for \emph{meta-training}.
%
We consider a model $f_{\btheta}$ parameterized by $\btheta$, which is updated iteratively.
Here, an updating~\emph{episode} includes an inner gradient update process (Line 8--Line 12)
and a meta-update process (Line 14).
Unlike MAML, for each task, additional adversarial samples are generated and used to enhance the robustness for meta-training.
Note that our algorithm is not restricted to any particular adversarial sample
generation method. We used the \emph{Fast Gradient Sign Method} (FGSM)~\cite{Goodfellow2015} in our experiments.
%
For task $\mT_i$, given a clean sample $({\bx_c}, {\by_c})$ from $\mD_i$, its corresponding adversarial sample $({\bx_{adv}}, {\by_{adv}})$ is generated using the following equations:
\begin{equation}
\begin{aligned}
{\bx_{adv}} & = {\bx_c} + \epsilon \mbox{sign}({\nabla}_{\bx_c} J(f_{{\btheta}_{pre}}, {\bx_c}, {\by_c})); \\
{\by_{adv}} & = {\by_c}.
\label{Eq:clean2adv}
\end{aligned}
\end{equation}
\noindent where $J(f_{{\btheta}_{pre}}, {\bx_c}, {\by_c})$ represents the cost used to train a classification model $f_{{\btheta}_{pre}}$ parameterized by ${\btheta}_{pre}$, and $\epsilon$ specifies the size of the adversarial perturbation (the larger the $\epsilon$, the higher the perturbation).
Note that the classification model $f_{{\btheta}_{pre}}$ is pre-trained based on the corresponding dataset and its parameters ${\btheta}_{pre}$ are fixed during the meta-training (Algorithm~\ref{Alg:ADML}) and meta-testing.

The key idea behind ADML is to utilize antagonistic correlations between clean and adversarial samples
to let the inner gradient update and the meta-update arm-wrestle with each other to obtain a good initialization
of model parameters $\btheta$, which is robust to adversarial samples.
Specifically, in the inner gradient update, we compute the new model parameters (updated in two directions) $\btheta_{adv_i}^{'}$ and $\btheta_{c_i}^{'}$
based on generated adversarial samples $\bD_{adv_i}$, 
and clean samples $\bD_{c_i}$ in training set $\bD_{i}$ of task $\mT_i$ respectively using gradient decent (Line 10).
%
%
%
%
In the meta-update process, we update the model parameters $\btheta$ by optimizing the losses  $\mL_{i}(f_{\btheta_{adv_i}^{'}})$ and $\mL_{i}(f_{\btheta_{c_i}^{'}})$ of the model with updated parameters $\btheta_{adv_i}^{'}$ and $\btheta_{c_i}^{'}$ with respect to ${\btheta}$
based on the clean samples $\bD_{c_{i}}^{'}$ in testing set $\bD'_{i}$ of task $\mT_i$ and the corresponding adversarial samples $\bD_{adv_{i}}^{'}$ respectively:
%
%
\begin{equation}
\begin{aligned}
\min \limits_{\btheta}\sum_{\mT_i \sim \mT}\mL_{i}(f_{\btheta_{adv_i}^{'}}, \bD_{c_{i}}^{'}) = \min \limits_{\btheta}\sum_{\mT_i \sim \mT}\mL_{i}(f_{\btheta - \alpha_1\nabla_{\btheta}\mL_{i}(f_{\btheta}, \bD_{adv_{i}})}, \bD_{c_{i}}^{'}); \\
\min \limits_{\btheta}\sum_{\mT_i \sim \mT}\mL_{i}(f_{\btheta_{c_i}^{'}}, \bD_{adv_{i}}^{'}) = \min \limits_{\btheta}\sum_{\mT_i \sim \mT}\mL_{i}(f_{\btheta - \alpha_2\nabla_{\btheta}\mL_{i}(f_{\btheta}, \bD_{c_{i}})}, \bD_{adv_{i}}^{'}).
\label{Eq:Meta-update}
\end{aligned}
\end{equation}
%

Note that in the meta-update, ${\btheta}$ is optimized in an \emph{adversarial} manner: the gradient of the loss of the model with $\btheta_{adv_i}^{'}$ (updated using adversarial samples $\bD_{adv_{i}}$) is calculated based on clean samples $\bD_{c_{i}}^{'}$, while the gradient of the loss of the model with $\btheta_{c_i}^{'}$ (updated using $\bD_{c_{i}}$) is calculated based on adversarial samples $\bD_{adv_{i}}^{'}$.
The arm-wrestling between the inner gradient update and the meta-update brings an obvious benefit: the model adapted to adversarial samples (through the inner gradient update using adversarial samples) is made suitable also for clean samples through the optimization of ${\btheta}$ in the meta-update based on the clean samples, and vice versa. So ``\emph{adversarial}'' in ADML refers to not only adversarial samples but also the way of meta-training.
%

\begin{algorithm*}[t]
\caption{Adversarial Meta-Learner (ADML)}
\label{Alg:ADML}
\begin{algorithmic}[1]
\STATE {\bf Require:} $\alpha_1/\alpha_2$ and $\beta_1/\beta_2$: The step sizes for inner gradient update and meta-update respectively \\
\STATE {\bf Require:} $\mD$: The datasets for meta-training \\
\STATE {\bf Require:} $<\mL_{i}(\cdot)>$: The loss function for task $\mT_i$, $\forall i \in \{1,\cdots,T\}$
\STATE Randomly initialize $\btheta$;
\WHILE {not done}
\STATE Sample batch of tasks $<\mT_i>$ from task set $\mT$;
\FORALL {$\mT_i$}
\STATE Sample $K$ clean samples $\left\{({\bx_c}^1, {\by_c}^1), ... ,({\bx_c}^{K}, {\by_c}^{K})\right\}$ from $\bD_i$;

\STATE Generate $K$ adversarial samples $\left\{({\bx_{adv}}^1, {\by_{adv}}^1), ... ,({\bx_{adv}}^{K}, {\by_{adv}}^{K})\right\}$ based on another $K$ clean samples from $\bD_i$
to form a dataset $\overline{\bD_{i}} := \left\{\bD_{adv_{i}}, \bD_{c_{i}}\right\}$ for the inner gradient update, containing $K$ adversarial samples and $K$ clean samples; \\



\STATE Compute updated model parameters with gradient descent respectively: \\
$\btheta_{adv_i}^{'} := \btheta - \alpha_1\nabla_{\btheta}\mL_{i}(f_{\btheta}, \bD_{adv_{i}})$;
$\btheta_{c_i}^{'} := \btheta - \alpha_2\nabla_{\btheta}\mL_{i}(f_{\btheta}, \bD_{c_{i}})$;

\STATE Sample $k$ clean samples $\left\{({\bx_c}^1, {\by_c}^1), ... ,({\bx_c}^{k}, {\by_c}^{k})\right\}$ from $\bD'_i$;

\STATE Generate $k$ adversarial samples $\left\{({\bx_{adv}}^1, {\by_{adv}}^1), ... ,({\bx_{adv}}^{k}, {\by_{adv}}^{k})\right\}$ based on another $k$ clean samples from $\bD'_i$
to form a dataset $\overline{\bD'_{i}} := \left\{\bD_{adv_{i}}^{'}, \bD_{c_{i}}^{'}\right\}$ for the meta-update, containing $k$ adversarial samples and $k$ clean samples; \\

\ENDFOR
\STATE Update $\btheta := \btheta - \beta_1\nabla_{\btheta}\sum_{\mT_i \sim \mT}\mL_{i}(f_{\btheta_{adv_i}^{'}}, \bD_{c_{i}}^{'})$; $\btheta := \btheta - \beta_2\nabla_{\btheta}\sum_{\mT_i \sim \mT}\mL_{i}(f_{\btheta_{c_i}^{'}}, \bD_{adv_{i}}^{'})$;
\ENDWHILE
\end{algorithmic}
\end{algorithm*}

%
%
%

%
The meta-update of the model parameters ${\btheta}$ is performed as the last step of each episode (Line 14).
Through the arm-wrestling between the inner gradient update and the meta-update in the meta-training process,
${\btheta}$ will be updated to a certain point, such that the average loss given by both adversarial samples
and clean samples of all the tasks is minimized. In addition, we set the step sizes $\alpha_1=\alpha_2=0.01, \beta_1=\beta_2=0.001$, and set $\mL_{i}(\cdot)$ of each classification task $\mT_i$ to be
the cross-entropy loss. $K$ and $k$ are task-specific, whose settings are discussed in the next section.
It can be easily seen that ADML preserves the model-agnostic property of MAML because both the inner gradient update and the meta-update processes are fully compatible with any learning model that can be trained by gradient descent.

\begin{figure}[t]
\centering
\includegraphics[scale=0.35]{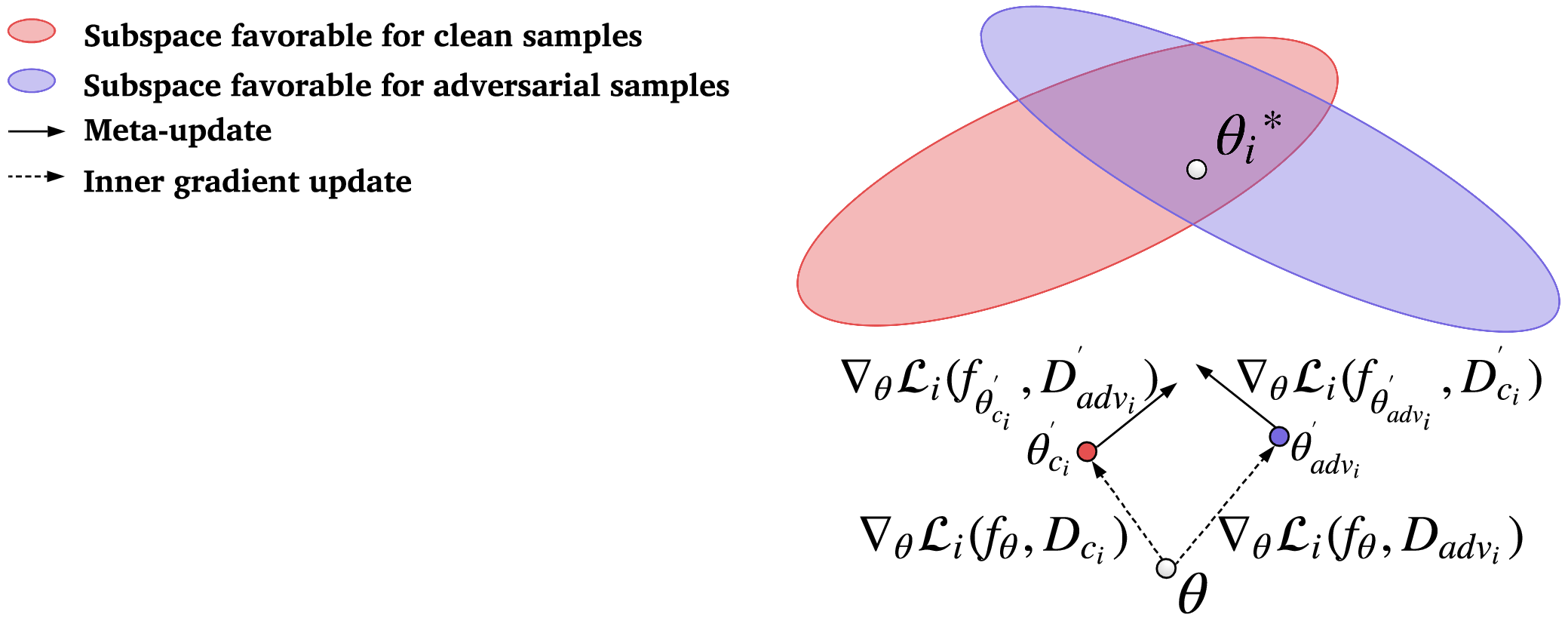}
\caption{Illustration of design philosophy of ADML}
\label{ADML}
\end{figure}

We further illustrate the design philosophy of our algorithm in Figure~\ref{ADML}.
For each task $\mT_i$, in the inner gradient update, ADML first drags ${\btheta}$ via gradient descent to the direction of the subspace that is favorable for adversarial samples (marked with the purple color) as well as another subspace that is favorable for clean samples (marked with the red color) to reach two points $\btheta_{adv_i}^{'}$ and $\btheta_{c_i}^{'}$ respectively (i.e., Line 10). Then in the meta-update, based on $\btheta_{adv_i}^{'}$ and $\btheta_{c_i}^{'}$, ADML further optimizes ${\btheta}$ to its antithetic subspaces respectively (i.e., Line 14), and hopefully ${\btheta}$ can reach the optimal point ${\btheta}_{i}^{*}$, which is supposed to fall into
the intersection of the subspace pair and is able to achieve a good trade-off between clean and adversarial samples to boost the overall performance on both samples.
Note that here we only show the updates via a single task.
Using all the tasks in $\mT$, ${\btheta}$ can be optimized to a point with the smallest average distance to the intersections of all the subspace pairs, and thus can be quickly adapted to new tasks even with adversarial samples.

As mentioned before, a rather straightforward solution to the above adversarial meta-learning problem is to simply
combine a meta-learner (e.g., MAML~\cite{finn2017model}) with adversarial training (e.g.,~\cite{Goodfellow2015}).
Specifically, we mix adversarial and clean samples to form both $\bD_i$ (used in the inner gradient update) and
$\bD'_i$ (used in the meta-update), which are then used to calculate $\btheta_{i}^{'}$ and update $\btheta$ using
the following equations (just like MAML) respectively:
\begin{equation}
\begin{aligned}
& \btheta_{i}^{'} = \btheta - \alpha\nabla_{\btheta}\mL_{i}(f_{\btheta}, \bD_{i}); \\
& \btheta \xleftarrow{} \btheta - \beta\nabla_{\btheta}\sum_{\mT_i \sim \mT}\mL_{i}(f_{{\btheta'_{i}}}, \bD'_{i}).
\label{Eq:theta0-theta}
\end{aligned}
\end{equation}



%
%
%
%

We call this method \emph{MAML-AD}, which is used as a baseline for performance evaluation.
However, it has been shown by our experimental results 
that although MAML-AD can slightly mitigate the problem, it still suffers from a significant performance degradation for new tasks with adversarial samples.
This clearly shows that simply involving adversarial samples during the meta-training does not necessarily enhance the model's robustness; 
and well justifies that our idea of doing the inner gradient update and the meta-update in an adversarial way is necessary.
\section{Performance Evaluation}
\label{Sec:Eval}
The goal of our evaluation is to test and verify three properties of ADML: 1) ADML can learn quickly from limited data via a few
gradient updates for a new task, and it is effective even in the cases with only clean samples;
2) ADML suffers from a minor performance degradation and yields much better performance than other meta-learning algorithms
when encountering adversarial samples; and 3) ADML maintains stable performance when the perturbation of adversarial samples (i.e., $\epsilon$)
escalates.
%
%
%
In this section, we first introduce the experimental setup, and then present and analyze the results.
\subsection{Experimental Setup}
\label{Sec:Setup}
In our experiments, we employed two commonly-used image benchmarks, MiniImageNet~\cite{vinyals2016matching}, and CIFAR100~\cite{krizhevsky2009learning}.
MiniImageNet is a benchmark for few-shot learning, which includes 100 classes and each of them has 600 samples.
CIFAR100
was created originally for object recognition tasks, whose data are suitable for meta-learning,
and just like MiniImageNet, it has 100 classes, each of which contains 600 images.
%
%
Similar as in~\cite{finn2017model}, we considered 1-shot and 5-shot 5-way classification tasks.
5 samples per class were used for the inner gradient update during meta-training of a 5-shot
learning model (one for 1-shot learning model). Thus $K$ in ADML was set to $25$ for 5-shot learning and $5$ for 1-shot learning.
15 samples per class were used for the meta-update, thus we set $k=75$.
During the meta-testing, the learning model was trained using samples of 5 unseen classes, then we tested it by using it to classify new instances into these 5 classes.
MiniImageNet was divided into 64, 16 and 20 classes for training, validation (for tuning hyperparamters) and testing respectively. We randomly sampled
5 classes from them to form each classification task.
Since CIFAR100 has not been used for meta-learning before, we created the meta-learning version of CIFAR100, which has the same settings
as MiniImageNet.
%
%

%
For FGSM~\cite{Goodfellow2015},
%
we leveraged a well-trained VGG16 network~\cite{Simonyan2014} for image classification (pre-trained on ImageNet~\cite{ILSVRC15} and CIFAR100 respectively) to generate adversarial samples.
%
The parameter $\epsilon$ 
was set to 2 when generating adversarial samples for the meta-training,
and was set to $2$ and $0.2$ for the meta-testing.
Note that while the FGSM is leveraged to generate adversarial samples,
the proposed ADML is agnostic to the particular choice of adversarial sample generation
method.
%
%
%
%

We compared ADML against three representative meta-learning algorithms, including MAML~\cite{finn2017model}, Matching Networks~\cite{vinyals2016matching}, and Relation Networks~\cite{sung2018learning}.
%
Moreover, for fair comparisons, we compared ADML with another adversarial meta-learner MAML-AD (introduced in the last section), which can be
considered as a rather straightforward extension of MAML.
For the implementation of ADML, we followed the architecture used by~\cite{finn2017model} for image embedding, which contains four $3 \times 3$ convolutional blocks with batch normalizations, ReLU activations and $2\times2$ max-poolings.
In our experiments, we used the implementation at~\cite{maml} for MAML,
the Full Contextual Embeddings (FCE)
implementation at~\cite{MatchingNetworks} for Matching Networks, and 
the implementation at~\cite{RelationNetworks} for Relation Networks.
%
%

%
\begin{table*}[t]\scriptsize
\centering
\renewcommand\arraystretch{1.3}
\caption{Average classification accuracies on MiniImageNet (5-way, 1-shot)}
\label{Mini-1shot}
\resizebox{\textwidth}{!}{
\begin{tabular}{|c|c|c|c|c|c|}
\hline
\multirow{2}{*}{Method} & \multirow{2}{*}{Meta-testing} & \multicolumn{2}{c|}{$\epsilon=2$} & \multicolumn{2}{c|}{$\epsilon=0.2$} \\ \cline{3-6} 
 &  & Clean & Adversarial & Clean & Adversarial \\ \hline
\multirow{2}{*}{MAML~\cite{finn2017model}} & Clean & $48.47\pm1.78\%$ & $28.63\pm1.54\%$ & $48.47\pm1.78\%$ & $42.13\pm1.75\%$ \\ \cline{2-6} 
 & Adversarial & $28.93\pm1.62\%$ & $30.73\pm1.66\%$ & $42.23\pm1.85\%$ & $40.17\pm1.76\%$ \\ \hline
\multirow{2}{*}{MAML-AD} & Clean & $43.13\pm1.88\%$ & $32.33\pm1.74\%$ & $43.13\pm1.88\%$ & $36.80\pm1.76\%$ \\ \cline{2-6} 
 & Adversarial & $32.47\pm1.60\%$ & $37.87\pm1.74\%$ & $37.63\pm1.64\%$ & $37.13\pm1.75\%$ \\ \hline
\multirow{2}{*}{Matching Nets~\cite{vinyals2016matching}} & Clean & $43.87\pm0.41\%$ & $30.02\pm0.39\%$ & $43.88\pm0.48\%$ & $36.14\pm0.40\%$ \\ \cline{2-6} 
 & Adversarial & $30.45\pm0.44\%$ & $30.80\pm0.43\%$ & $36.58\pm0.49\%$ & $35.03\pm0.39\%$ \\ \hline
\multirow{2}{*}{Relation Nets~\cite{sung2018learning}} & Clean & \boldmath{$49.67\pm0.85\%$} & $32.32\pm0.58\%$ & \boldmath{$49.45\pm0.84\%$} & $43.03\pm0.74\%$ \\ \cline{2-6} 
 & Adversarial & $32.59\pm0.79\%$ & $32.85\pm0.63\%$ & $42.98\pm0.85\%$ & $40.89\pm0.79\%$ \\ \hline
\multirow{2}{*}{ADML (Ours)} & Clean & $48.00\pm1.87\%$ & \boldmath{$43.00\pm1.88\%$} & $48.00\pm1.87\%$ & \boldmath{$43.20\pm1.70\%$} \\ \cline{2-6} 
& Adversarial & \boldmath{$40.10\pm1.73\%$} & \boldmath{$40.70\pm1.74\%$} & \boldmath{$44.00\pm1.83\%$} & \boldmath{$41.20\pm1.75\%$} \\ \hline
\end{tabular}
}
\vspace{-6mm}
\end{table*}
%
\subsection{Experimental Results}
\label{Sec:Results}
To fully test the effectiveness of ADML, we conducted
a comprehensive empirical study, which covers various possible cases.
The experimental results
on MiniImageNet and CIFAR100 are presented in
Tables~\ref{Mini-1shot}--\ref{Mini-5shot} and Tables~\ref{CIFAR-1shot}--\ref{CIFAR-5shot} (in supplementary materials) respectively.
%
Each entry in these tables gives the average classification accuracy (with $95\%$ confidence intervals) of the corresponding test case,
and the best results for each test case are marked in bold.

The experiments were conducted in six different test cases (combinations):
``\emph{Clean-Clean}'', ``\emph{Clean-Adversarial}'', ``\emph{Adversarial-Clean}'', ``\emph{Adversarial-Adversarial}'', ``\emph{$40\%$-Clean}'' and ``\emph{$40\%$-Adversarial}''.
The first part of each combination (corresponding to a row) represents the training data used in the inner gradient update (or support set for Matching Networks and Relation Networks) 
during the meta-testing,
while the second part (corresponding to a column) represents the testing data (or query set for Matching Networks and Relation Networks) for evaluation.
``Clean'' means clean samples only; ``Adversarial'' means adversarial samples only; 
and ``$40\%$'' means that $40\%$ samples of each class are adversarial and the rest $60\%$ are
clean, which represents intermediate cases. 
Note that the combinations, ``$40\%$-Clean'' and ``$40\%$-Adversarial'', do not exist for 1-shot learning since there is only one sample per class.
Based on the results in Tables~\ref{Mini-1shot}--\ref{Mini-5shot}, we can make the following observations:

%
%
%
\begin{table*}[t]\scriptsize
\centering
\renewcommand\arraystretch{1.3}
\caption{Average classification accuracies on MiniImageNet (5-way, 5-shot)}
\label{Mini-5shot}
\resizebox{\textwidth}{!}{
\begin{tabular}{|c|c|c|c|c|c|}
\hline
\multirow{2}{*}{Method} & \multirow{2}{*}{Meta-testing} & \multicolumn{2}{c|}{$\epsilon=2$} & \multicolumn{2}{c|}{$\epsilon=0.2$} \\ \cline{3-6} 
 &  & Clean & Adversarial & Clean & Adversarial \\ \hline
\multirow{3}{*}{MAML~\cite{finn2017model}} & Clean & $61.45\pm0.91\%$ & $36.65\pm0.88\%$ & $61.47\pm0.91\%$ & $53.05\pm0.86\%$ \\ \cline{2-6} 
 & $40\%$ & $56.74\pm0.93\%$ & $43.05\pm0.86\%$ & $59.25\pm0.91\%$ & $54.67\pm0.91\%$ \\ \cline{2-6} 
 & Adversarial & $41.49\pm0.95\%$ & $45.46\pm0.97\%$ & $55.19\pm0.95\%$ & $53.33\pm0.92\%$ \\ \hline
\multirow{3}{*}{MAML-AD} & Clean & $57.13\pm0.96\%$ & $41.65\pm0.92\%$ & $57.09\pm0.96\%$ & $49.71\pm0.88\%$ \\ \cline{2-6} 
 & $40\%$ & $54.07\pm0.91\%$ & $48.74\pm0.91\%$ & $56.52\pm0.90\%$ & $52.08\pm0.90\%$ \\ \cline{2-6} 
 & Adversarial & $43.21\pm0.91\%$ & $52.07\pm0.96\%$ & $51.23\pm0.90\%$ & $51.36\pm0.94\%$ \\ \hline
\multirow{3}{*}{Matching Nets~\cite{vinyals2016matching}} & Clean & $55.99\pm0.47\%$ & $33.73\pm0.39\%$ & $55.55\pm0.44\%$ & $44.91\pm0.40\%$ \\ \cline{2-6} 
 & $40\%$ & $49.88\pm0.45\%$ & $35.67\pm0.44\%$ & $52.72\pm0.45\%$ & $45.65\pm0.42\%$ \\ \cline{2-6} 
 & Adversarial & $36.24\pm0.45\%$ & $37.91\pm0.40\%$ & $47.77\pm0.45\%$ & $46.19\pm0.44\%$ \\ \hline
\multirow{3}{*}{Relation Nets~\cite{sung2018learning}} & Clean & \boldmath{$63.85\pm0.73\%$} & $38.37\pm0.64\%$ & \boldmath{$63.86\pm0.73\%$} & $55.39\pm0.68\%$ \\ \cline{2-6} 
 & $40\%$ & $56.53\pm0.77\%$ & $41.04\pm0.67\%$ & $59.02\pm0.70\%$ & $55.06\pm0.68\%$ \\ \cline{2-6} 
 & Adversarial & $42.74\pm0.79\%$ & $46.08\pm0.63\%$ & $56.85\pm0.73\%$ & $53.65\pm0.69\%$ \\ \hline
\multirow{3}{*}{ADML (Ours)} & Clean & $59.38\pm0.99\%$ & \boldmath{$57.03\pm0.98\%$} & $59.40\pm0.99\%$ & \boldmath{$56.07\pm0.96\%$} \\ \cline{2-6} 
 & $40\%$ & \boldmath{$58.12\pm0.90\%$} & \boldmath{$55.22\pm0.98\%$} & \boldmath{$59.67\pm0.89\%$} & \boldmath{$56.49\pm0.92\%$} \\ \cline{2-6} 
 & Adversarial & \boldmath{$58.06\pm0.96\%$} & \boldmath{$55.27\pm0.92\%$} & \boldmath{$57.44\pm0.88\%$} & \boldmath{$54.47\pm0.93\%$} \\ \hline
\end{tabular}
}
\vspace{-6mm}
\end{table*}
1) Just like MAML, ADML is indeed an effective meta-learner since it leads to quick learning from a small amount of new data for a new task.
In the ``Clean-Clean'' cases, the general condition of meta-learning,
ADML delivers desirable results, which are very close to the state-of-the-art given
by MAML and Relation Networks, and consistently better than those of MAML-AD and Matching Networks.
For example, in the case of 5-way 1-shot classification with $\epsilon=2$ (Table~\ref{Mini-1shot}),
ADML gives an average classification accuracy of $48.00\%$, which is very close to that given by MAML (i.e., $48.47\%$) and Relation Networks ($49.67\%$),
and it performs better than MAML-AD ($43.13\%$) and Matching Networks ($43.87\%$).
\emph{Note that the proposed ADML focuses on the cases with adversarial samples, and it is reasonable that it performs slightly worse on the cases with only clean samples.}
%
%
%
%
%
%

2) ADML is robust to adversarial samples since it only suffers from a minor performance degradation when encountering adversarial samples. For example, for the 5-way 5-shot classification with $\epsilon=2$ (Table~\ref{Mini-5shot}),
ADML gives classification accuracies of $57.03\%$, $58.06\%$, $55.27\%$, $58.12\%$ and $55.22\%$ in the five test cases respectively.
%
%
Compared to the ``Clean-Clean'' case (i.e., $59.38\%$),
the performance degradation is only $4.16\%$ in the worst-case and $2.64\%$ on average.
%
However, the classification accuracies given by the other meta-learning algorithms, including MAML-AD, drop substantially when there are adversarial samples.
For example, 
for MAML, the accuracy drops from $61.45\%$ (``Clean-Clean'') to $36.65\%$ (``Clean-Adversarial'') when injecting adversarial samples for testing, which represents a significant degradation of $24.80\%$.
%
%
Similar observations can be made for MAML-AD, Matching Networks and Relation Networks, which represent substantial degradations of $15.48\%$, $22.26\%$ and $25.48\%$ respectively.
%
%
%
%

3) ADML consistently outperforms all the other meta-learning algorithms in the test cases with adversarial samples. 
%
%
For instance, 
%
in the ``Clean-Adversarial'' cases of 5-way 5-shot learning with $\epsilon=2$ (Table~\ref{Mini-5shot}), 
ADML achieves an accuracy of $57.03\%$, which represents $20.38\%$, $15.38\%$, $23.30\%$ and $18.66\%$ improvements over MAML, MAML-AD, Matching Networks and Relation Networks respectively.
%
%
%
%
This clearly shows the superiority of the adversarial meta-training procedure of the proposed ADML compared to MAML-AD, the straightforward adversarial meta-learner.
Even so, MAML-AD still generally performs better than the rest meta-training algorithms, when dealing with adversarial samples.
%
%

4) When the perturbation of adversarial samples escalates, ADML maintains stable performance.
For example, for 5-way 1-shot learning, when $\epsilon$ increases from $0.2$ to $2$ (Table~\ref{Mini-1shot}), ADML only leads to minor degradations of $0.2\%$, $3.9\%$ and $0.5\%$
in the corresponding three cases involving adversarial samples. 
However, much more significant degradations can be observed when the other meta-learning algorithms are applied.
For instance, the accuracies of MAML suffer from $13.50\%$, $13.30\%$ and $9.44\%$ drops in these three cases when increasing $\epsilon$ from $0.2$ to $2$.
%
%
\begin{figure*}[ht!]
    \centering
    \begin{subfigure}[t]{0.25\textwidth}
        \centering
        \includegraphics[width=\textwidth]{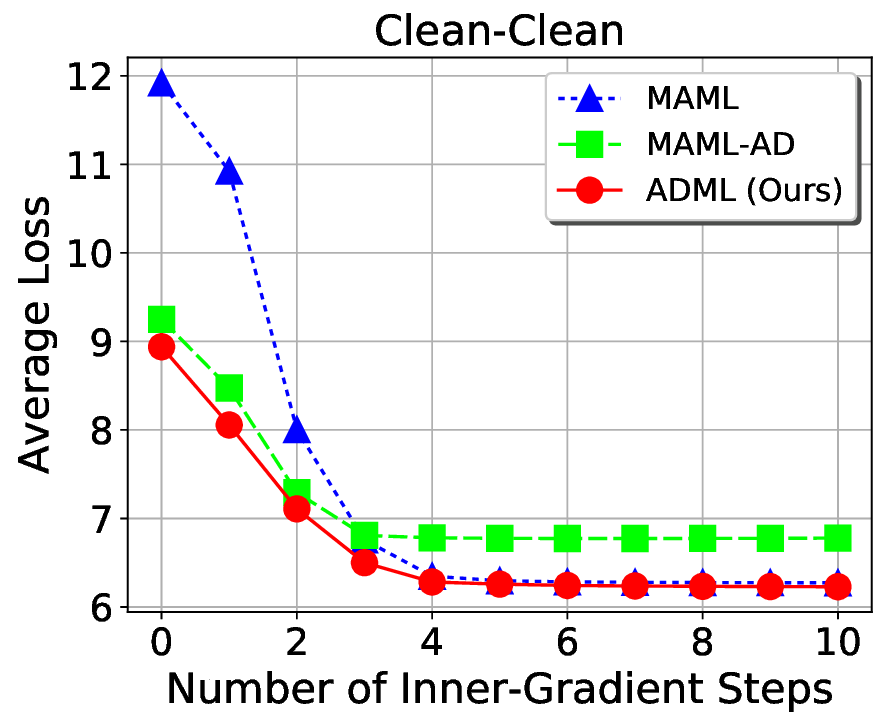}
    \end{subfigure}
    \hspace{-0.025\textwidth}
    ~ 
    \begin{subfigure}[t]{0.25\textwidth}
        \centering
        \includegraphics[width=\textwidth]{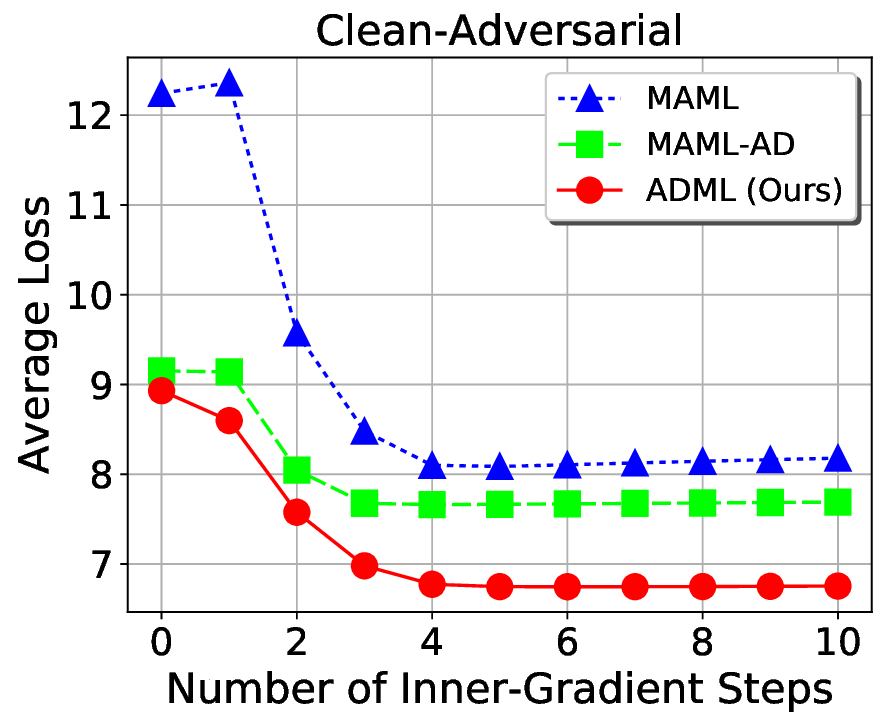}
    \end{subfigure}
    \hspace{-0.025\textwidth}
    ~ 
    \begin{subfigure}[t]{0.25\textwidth}
        \centering
        \includegraphics[width=\textwidth]{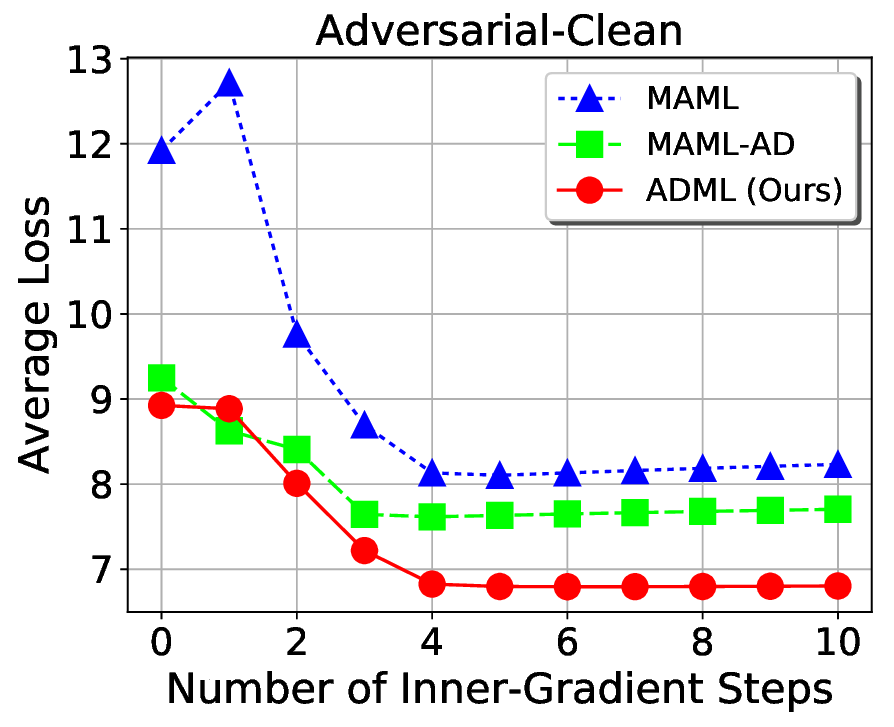}
    \end{subfigure}
    \hspace{-0.025\textwidth}
    ~
    \begin{subfigure}[t]{0.25\textwidth}
        \centering
        \includegraphics[width=\textwidth]{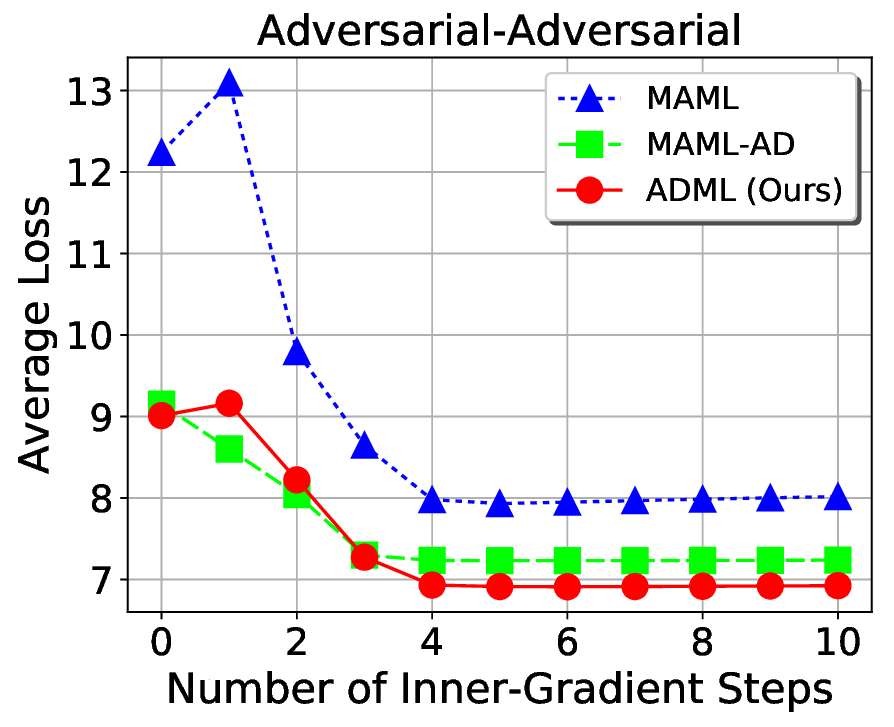}
    \end{subfigure}
    \caption{Average loss over the gradient update step for 5-way 1-shot learning on MiniImageNet}
    \label{Average-Loss-1}
    \vspace{-4mm}
\end{figure*}
\begin{figure*}[ht!]
    \centering
    \begin{subfigure}[t]{0.25\textwidth}
        \centering
        \includegraphics[width=\textwidth]{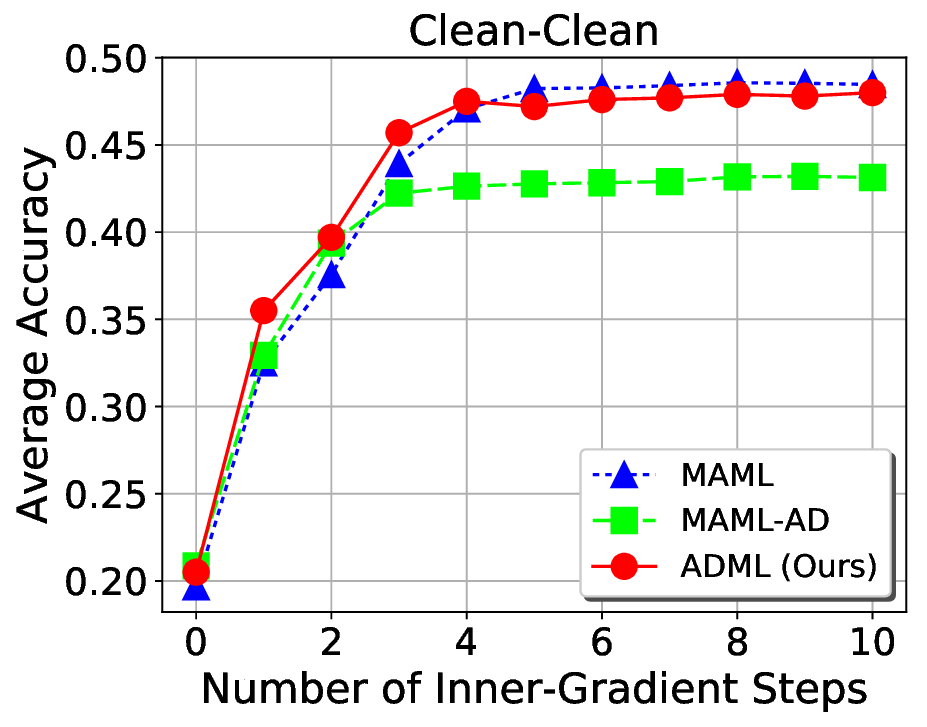}
    \end{subfigure}
    \hspace{-0.025\textwidth}
    ~ 
    \begin{subfigure}[t]{0.25\textwidth}
        \centering
        \includegraphics[width=\textwidth]{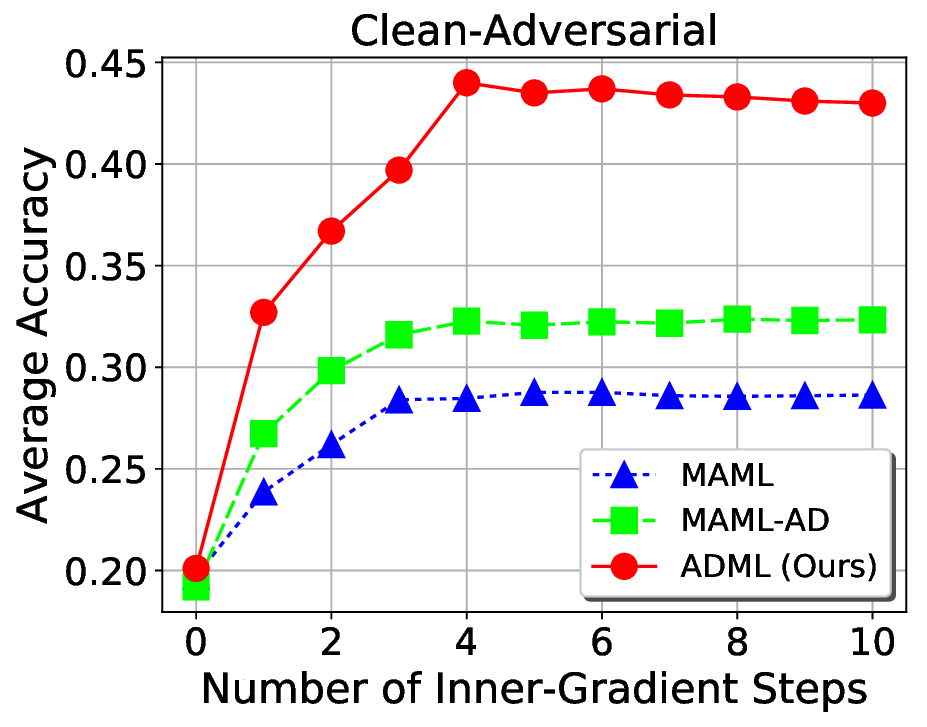}
    \end{subfigure}
    \hspace{-0.025\textwidth}
    ~ 
    \begin{subfigure}[t]{0.25\textwidth}
        \centering
        \includegraphics[width=\textwidth]{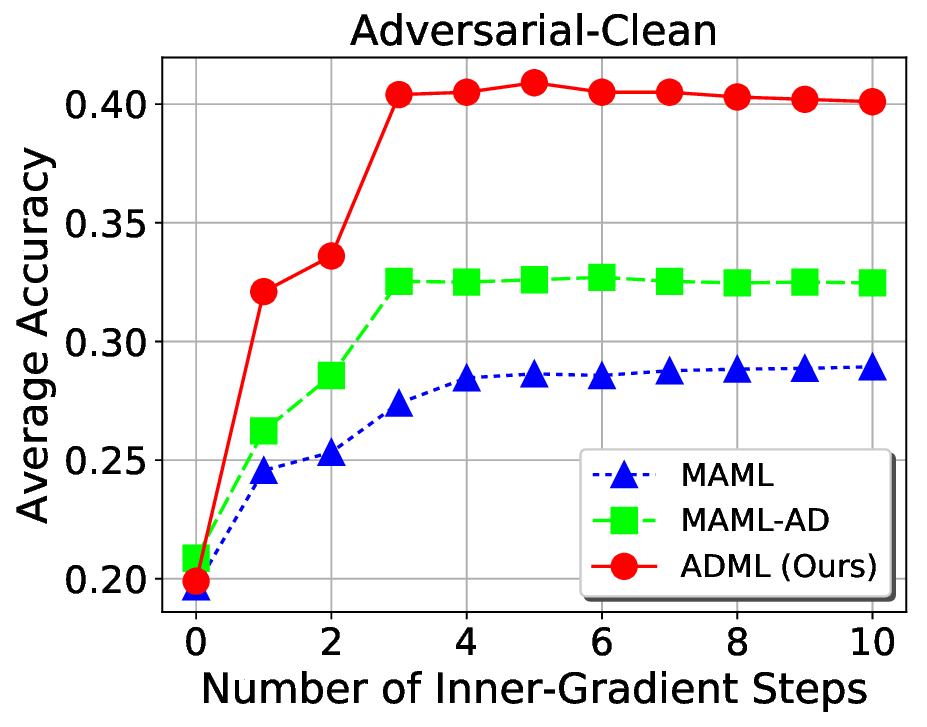}
    \end{subfigure}
    \hspace{-0.025\textwidth}
    ~
    \begin{subfigure}[t]{0.25\textwidth}
        \centering
        \includegraphics[width=\textwidth]{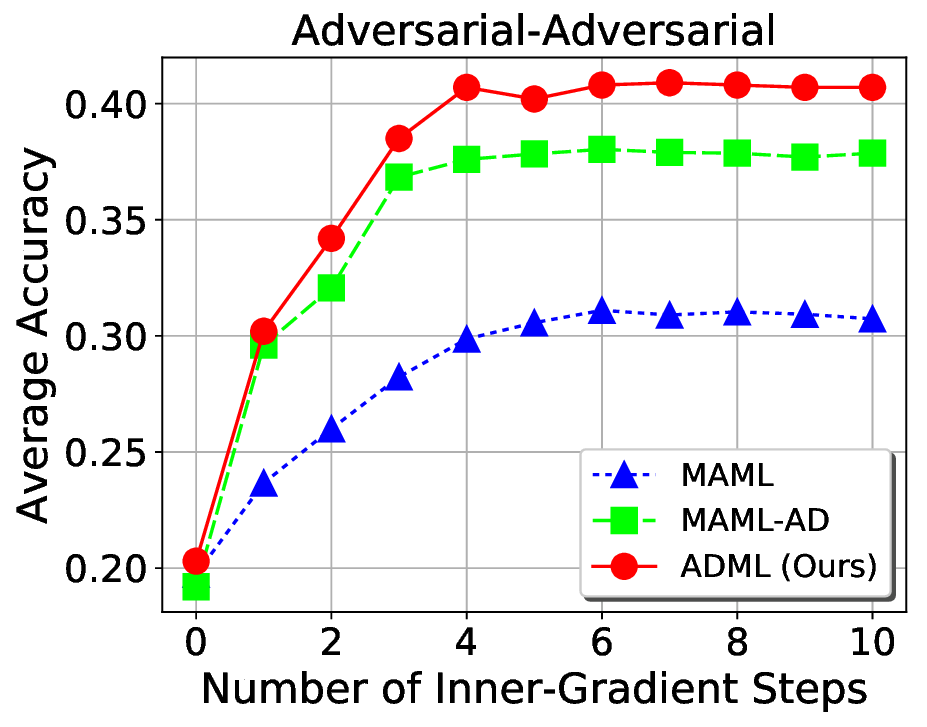}
    \end{subfigure}
    \caption{Top-1 accuracy over the gradient update step for 5-way 1-shot learning on MiniImageNet}
    \label{Top1-Acc-1}
    \vspace{-4mm}
\end{figure*}

5) As we might expect, when lower perturbations exerted, the adversarial samples are not significantly different from the corresponding clean samples, which brings about relatively close results in different test cases, not only for ADML, but also for the other meta-learning algorithms.
For instance, for the 5-way 5-shot classification with $\epsilon=0.2$ (Table~\ref{Mini-5shot}), MAML gives a smaller gap of $8.42\%$ between ``Clean-Clean'' and ``Clean-Adversarial'', compared with that of $24.80\%$ when $\epsilon=2$.

6) As expected, the classification accuracy increases dramatically when going from 1-shot to 5-shot learning. Particularly, when $\epsilon=2$, if we do 5-shot learning with ADML, we can achieve classification accuracies of $59.38\%$, $57.03\%$, $58.06\%$ and $55.27\%$ in the corresponding four test cases respectively, which represent $11.38\%$, $14.03\%$, $17.96\%$ and $14.57\%$ improvements over 1-shot learning. 
This observation implies that more training samples (even if they may be adversarial samples) lead to better classification accuracies.
%

%
%
%

\emph{Similar observations can be made for the results corresponding to 
CIFAR100 (i.e., Tables~\ref{CIFAR-1shot}--\ref{CIFAR-5shot} in supplementary materials).}
In addition, we show how the loss and Top-1 accuracy change in Figures~\ref{Average-Loss-1} and~\ref{Top1-Acc-1} during the meta-testing.  
%
Specifically, these two figures show that when ADML, MAML and MAML-AD are applied on MiniImageNet, how the losses and Top-1 accuracies change with the gradient update step
during the meta-testing in the four test cases 
of 5-way 1-shot learning with $\epsilon=2$.
We observe that, for all the cases,
the losses of the models learned with ADML drop sharply after only several gradient updates, and stabilize
at small values during the meta-testing, which are generally lower than those of the other two methods. 
Moreover, the Top-1 accuracies of the models learned with ADML rise sharply after only several gradient updates, and stabilize at values, which are generally higher than those of the other two methods (a little bit lower than that of MAML in ``Clean-Clean'' case).
\emph{Similar trends can be observed in Figures~\ref{Average-Loss-5}--\ref{Top1-Acc-5} in supplementary materials for 5-way 5-shot learning with $\epsilon=2$.}
These observations further confirm that ADML is suitable for meta-learning since it can
quickly learn and adapt from small data for a new task through only a few gradient updates.
%
%

\section{Conclusions}
\label{Sec:Conclusions}
In this paper, we proposed a novel 
method called ADML (ADversarial Meta-Learner) for meta-learning with adversarial samples, which features an \emph{adversarial} way for optimizing model parameters ${\btheta}$ during meta-training through the arm-wrestling between inner gradient update and meta-update using both clean and adversarial samples.
A comprehensive empirical study has been conducted for performance evaluation
using two widely-used datasets, MiniImageNet and CIFAR100.
The extensive experimental results have showed that 1) ADML is an effective meta-learner even in the cases with only clean samples;
2) a straightforward adversarial meta-learner, namely, MAML-AD, does not work well with adversarial samples;
%
in addition,
3) ADML is robust to adversarial samples and 
outperforms other meta-learning algorithms including MAML on adversarial meta-learning tasks;
and most importantly,
4) it opens up an interesting research direction and sheds light on dealing with the difficult cases with limited and even contaminated samples.
%
\section*{Broader Impact}
\label{Sec:Impact}
%
%
This work presents a novel method called ADML (ADversarial Meta-Learner) for meta-learning to deal with the cases involving adversarial samples.
%
%
The proposed algorithm sheds light on tackling the cases with limited and even contaminated samples, which are challenging but common in real life.
For example, for some rare events, there may be only a small number of photos taken in a bad environment (such as mist, rain, etc)
which may pose challenges on the learning ability of a model.
The proposed ADML is well suited for such cases.
Meanwhile, ADML may have some potential negative impact.
%
%
%
The failure of ADML can cause incorrect classifications, which may lead to false alarms or wrong decision-makings
that could further result in extra and unnecessary labor work.
In addition, our task/method does not leverage biases in the data. 

\bibliographystyle{unsrt}
\bibliography{Meta-Adversarial-NeurIPS20.bib}

\newpage

%
\begin{figure*}[t]
    \centering
    \begin{subfigure}[t]{0.25\textwidth}
        \centering
        \includegraphics[width=\textwidth]{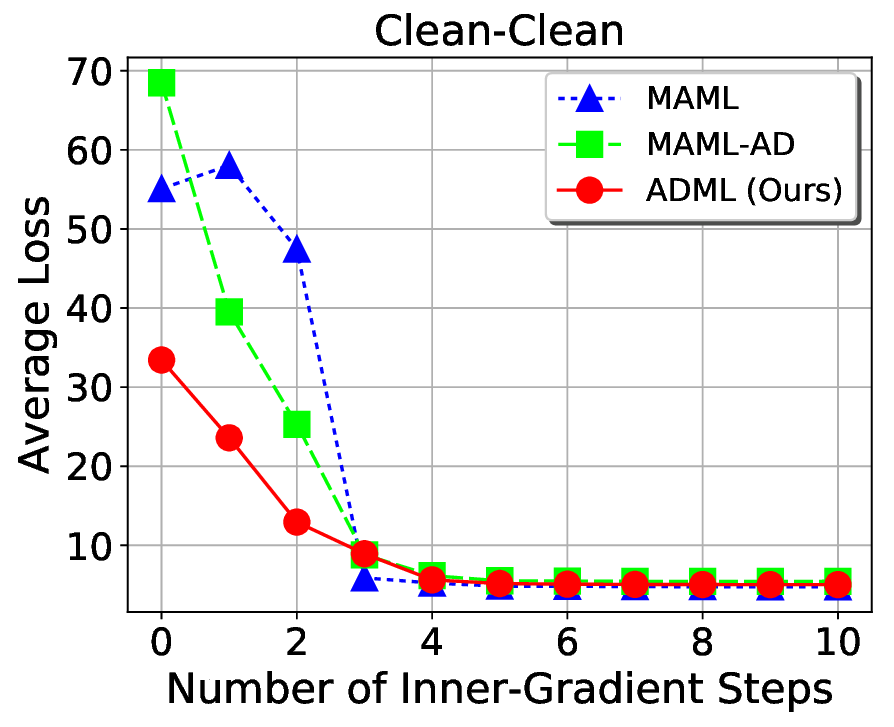}
    \end{subfigure}
    \hspace{-0.025\textwidth}
    ~ 
    \begin{subfigure}[t]{0.25\textwidth}
        \centering
        \includegraphics[width=\textwidth]{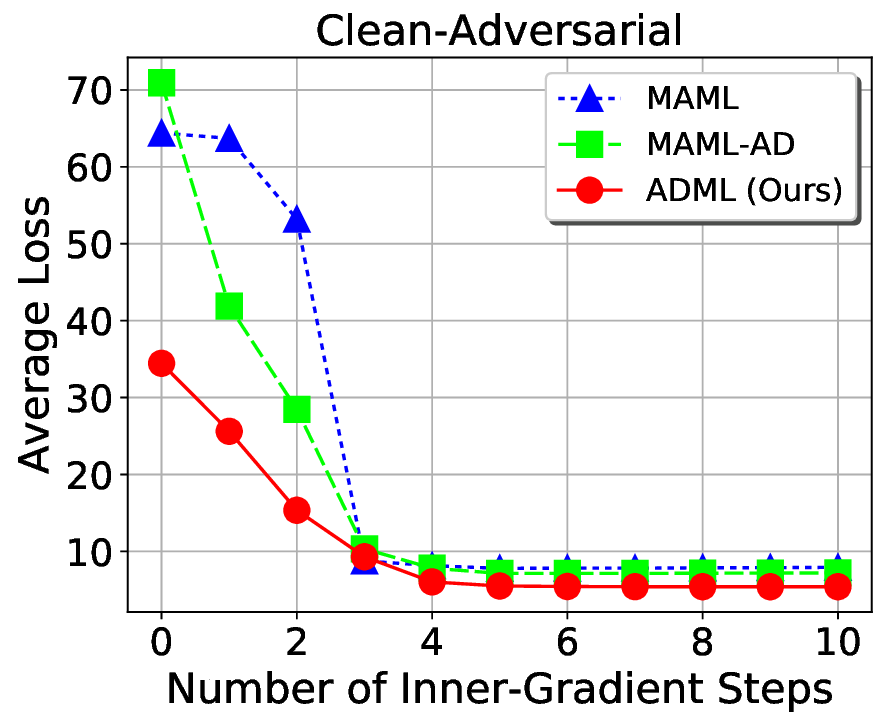}
    \end{subfigure}
    \hspace{-0.025\textwidth}
    ~ 
    \begin{subfigure}[t]{0.25\textwidth}
        \centering
        \includegraphics[width=\textwidth]{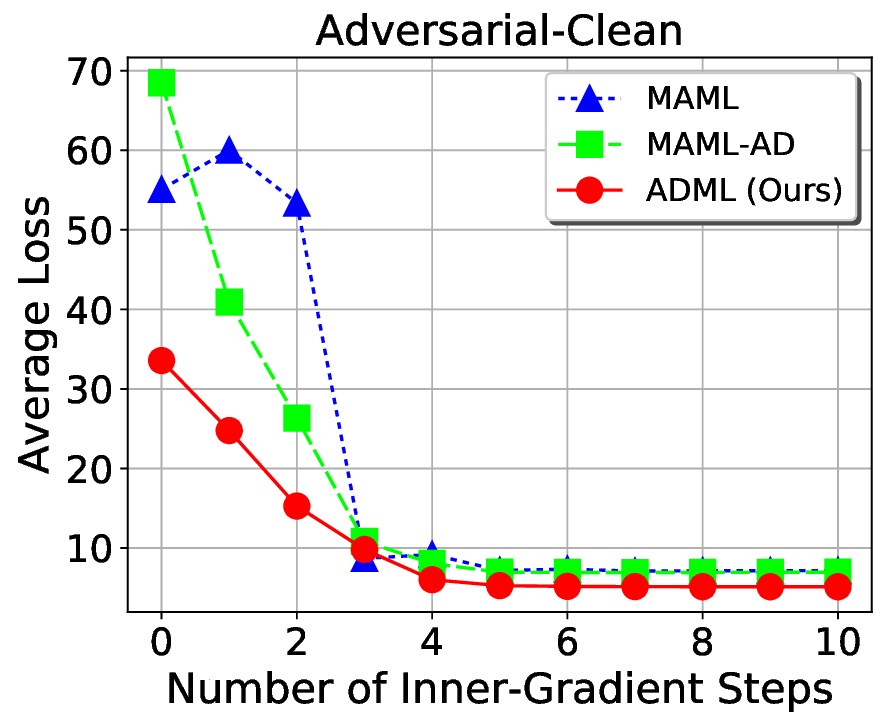}
    \end{subfigure}
    \hspace{-0.025\textwidth}
    ~
    \begin{subfigure}[t]{0.25\textwidth}
        \centering
        \includegraphics[width=\textwidth]{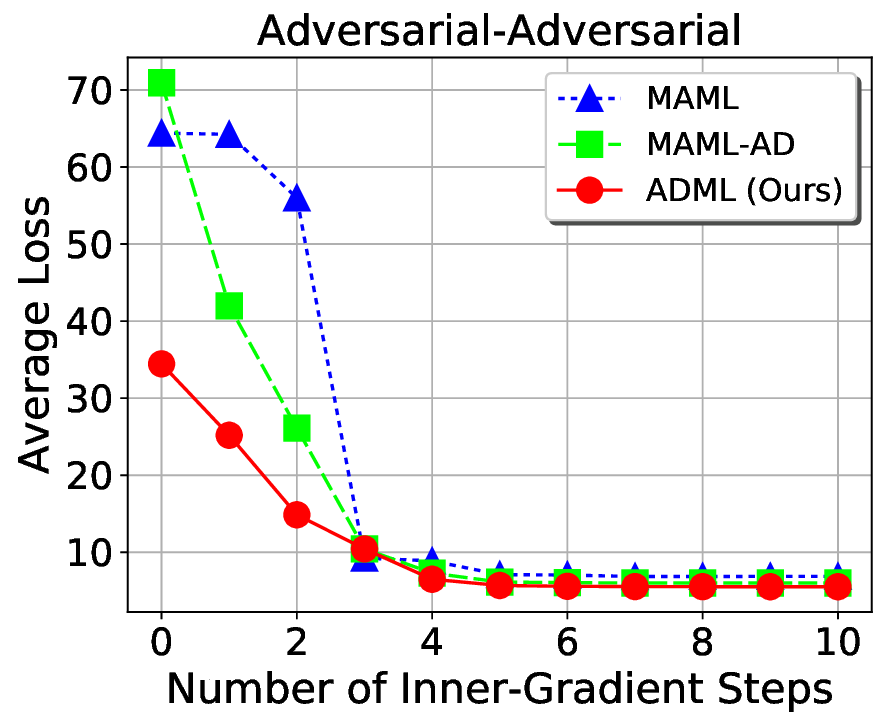}
    \end{subfigure}
    \caption{Average loss over the gradient update step for 5-way 5-shot learning on MiniImageNet}
    \label{Average-Loss-5}
\end{figure*}
\begin{figure*}[t]
    \centering
    \begin{subfigure}[t]{0.25\textwidth}
        \centering
        \includegraphics[width=\textwidth]{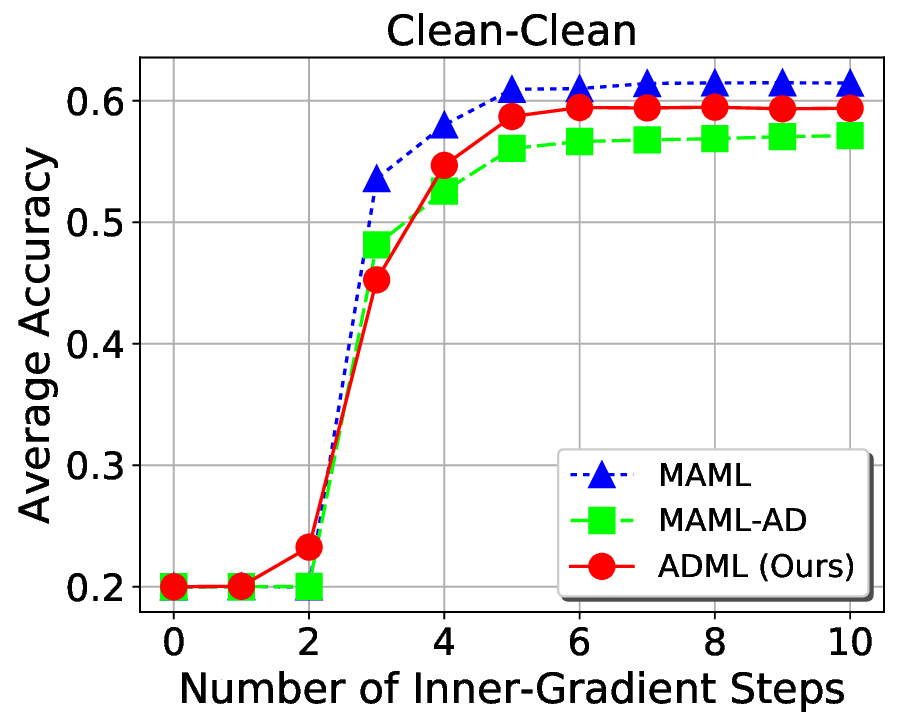}
    \end{subfigure}
    \hspace{-0.025\textwidth}
    ~ 
    \begin{subfigure}[t]{0.25\textwidth}
        \centering
        \includegraphics[width=\textwidth]{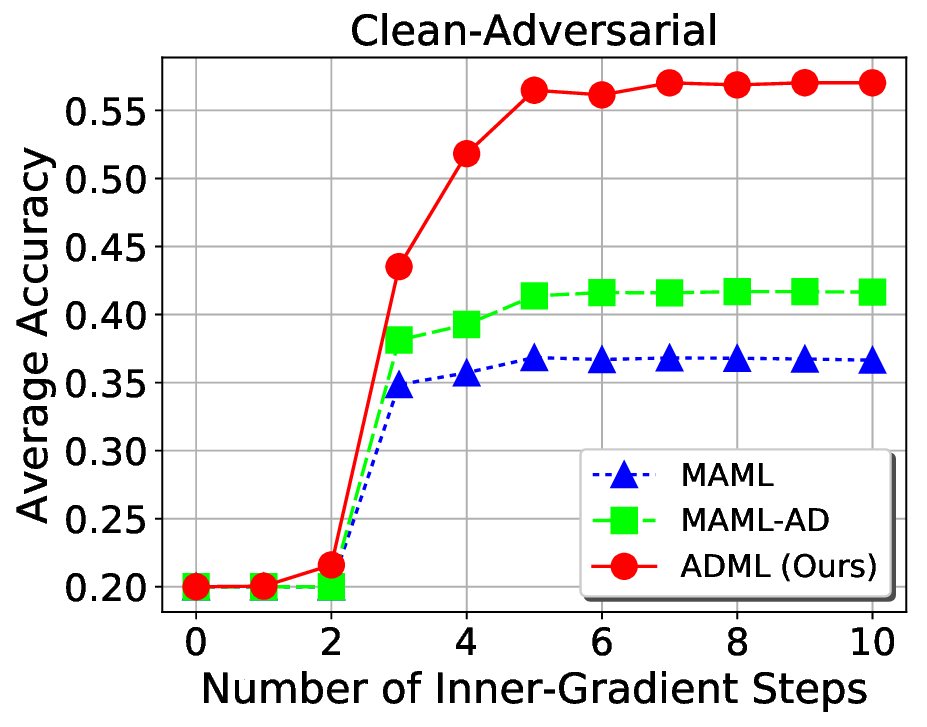}
    \end{subfigure}
    \hspace{-0.025\textwidth}
    ~ 
    \begin{subfigure}[t]{0.25\textwidth}
        \centering
        \includegraphics[width=\textwidth]{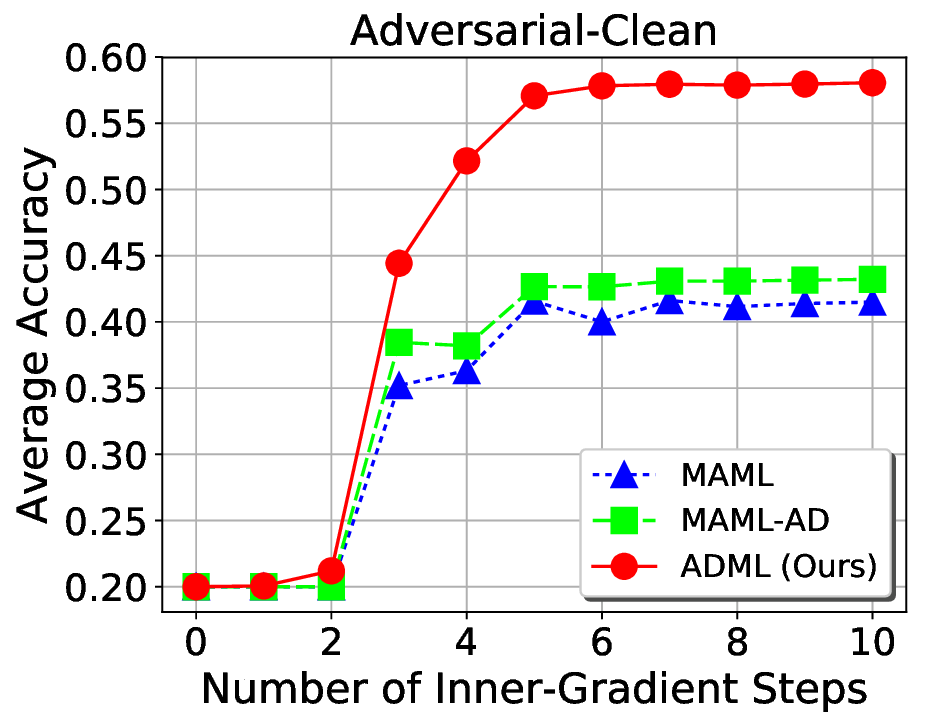}
    \end{subfigure}
    \hspace{-0.025\textwidth}
    ~
    \begin{subfigure}[t]{0.25\textwidth}
        \centering
        \includegraphics[width=\textwidth]{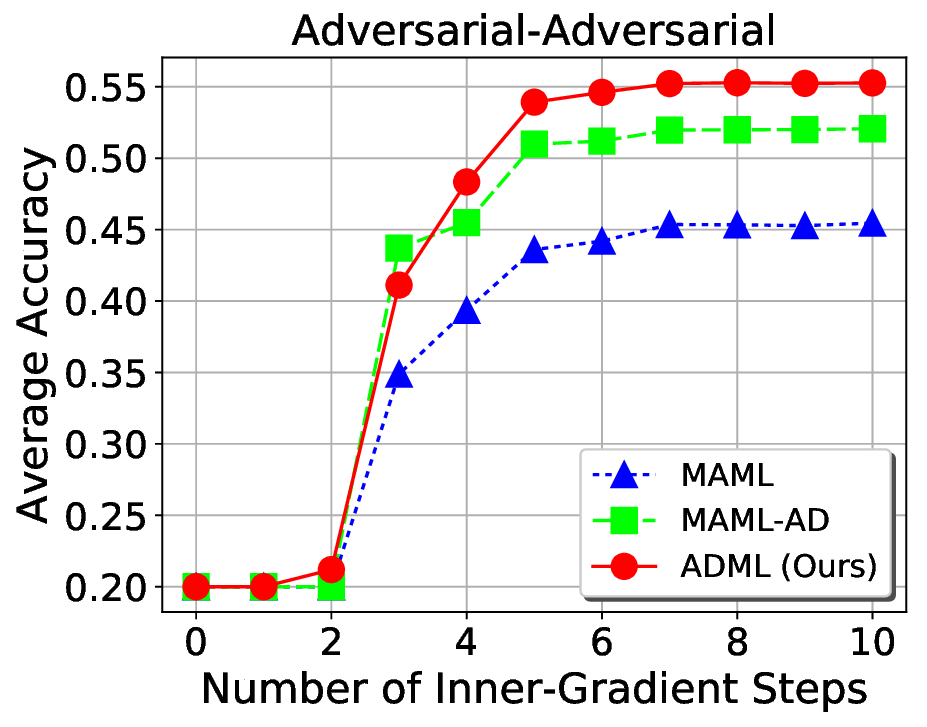}
    \end{subfigure}
    \caption{Top-1 accuracy over the gradient update step for 5-way 5-shot learning on MiniImageNet}
    \label{Top1-Acc-5}
\end{figure*}
\begin{table*}[t]\scriptsize
\setlength{\abovecaptionskip}{0.cm}
\setlength{\belowcaptionskip}{-0.cm}
\centering
\renewcommand\arraystretch{1.3}
\caption{Average classification accuracies on CIFAR100 (5-way, 1-shot)}
\label{CIFAR-1shot}
\resizebox{\textwidth}{!}{
\begin{tabular}{|c|c|c|c|c|c|}
\hline
\multirow{2}{*}{Method} & \multirow{2}{*}{Meta-testing} & \multicolumn{2}{c|}{$\epsilon=2$} & \multicolumn{2}{c|}{$\epsilon=0.2$} \\ \cline{3-6} 
 &  & Clean & Adversarial & Clean & Adversarial \\ \hline
\multirow{2}{*}{MAML~\cite{finn2017model}} & Clean & $57.67\pm1.76\%$ & $26.40\pm1.55\%$ & $57.67\pm1.76\%$ & $43.30\pm1.68\%$ \\ \cline{2-6} 
 & Adversarial & $28.13\pm1.56\%$ & $28.23\pm1.64\%$ & $43.03\pm1.76\%$ & $39.00\pm1.70\%$ \\ \hline
\multirow{2}{*}{MAML-AD} & Clean & $52.70\pm1.89\%$ & $36.20\pm1.65\%$ & $52.70\pm1.89\%$ & $39.17\pm1.82\%$ \\ \cline{2-6} 
 & Adversarial & $37.27\pm1.72\%$ & $41.67\pm1.86\%$ & $37.80\pm1.70\%$ & $37.60\pm1.78\%$ \\ \hline
\multirow{2}{*}{Matching Nets~\cite{vinyals2016matching}} & Clean & $47.94\pm0.56\%$ & $25.06\pm0.36\%$ & $47.68\pm0.52\%$ & $39.03\pm0.51\%$ \\ \cline{2-6} 
 & Adversarial & $24.82\pm0.46\%$ & $27.72\pm0.43\%$ & $40.08\pm0.57\%$ & $37.79\pm0.44\%$ \\ \hline
\multirow{2}{*}{Relation Nets~\cite{sung2018learning}} & Clean & \boldmath{$58.68\pm0.92\%$} & $31.11\pm0.93\%$ & \boldmath{$58.72\pm0.90\%$} & $45.03\pm0.76\%$ \\ \cline{2-6} 
 & Adversarial & $30.85\pm0.92\%$ & $30.52\pm0.59\%$ & $45.85\pm1.01\%$ & $41.40\pm0.80\%$ \\ \hline
\multirow{2}{*}{ADML (Ours)} & Clean & $55.70\pm2.00\%$ & \boldmath{$50.90\pm1.84\%$} & $55.70\pm2.00\%$ & \boldmath{$49.30\pm1.76\%$} \\ \cline{2-6} 
 & Adversarial & \boldmath{$54.50\pm1.69\%$} & \boldmath{$50.60\pm1.83\%$} & \boldmath{$52.90\pm1.92\%$} & \boldmath{$45.00\pm1.79\%$} \\ \hline
\end{tabular}
}
\end{table*}
\begin{table*}[t]\scriptsize
\setlength{\abovecaptionskip}{0.cm}
\setlength{\belowcaptionskip}{-0.cm}
\centering
\renewcommand\arraystretch{1.3}
\caption{Average classification accuracies on CIFAR100 (5-way, 5-shot)}
\label{CIFAR-5shot}
\resizebox{\textwidth}{!}{
\begin{tabular}{|c|c|c|c|c|c|}
\hline
\multirow{2}{*}{Method} & \multirow{2}{*}{Meta-testing} & \multicolumn{2}{c|}{$\epsilon=2$} & \multicolumn{2}{c|}{$\epsilon=0.2$} \\ \cline{3-6} 
 &  & Clean & Adversarial & Clean & Adversarial \\ \hline
\multirow{3}{*}{MAML~\cite{finn2017model}} & Clean & $74.03\pm0.89\%$ & $31.29\pm0.78$\% & $74.03\pm0.89\%$ & $54.15\pm1.00\%$ \\ \cline{2-6} 
 & $40\%$ & $65.69\pm0.92\%$ & $36.14\pm0.84\%$ & $68.99\pm0.94\%$ & $55.79\pm0.98\%$ \\ \cline{2-6} 
 & Adversarial & $33.34\pm0.90\%$ & $43.66\pm0.86\%$ & $59.08\pm1.00\%$ & $53.93\pm0.96\%$ \\ \hline
\multirow{3}{*}{MAML-AD} & Clean & $67.71\pm0.96\%$ & $44.61\pm0.90\%$ & $67.73\pm0.96\%$ & $56.07\pm0.95\%$ \\ \cline{2-6} 
 & $40\%$ & $64.85\pm0.90\%$ & $53.59\pm0.88\%$ & $65.93\pm0.93\%$ & $57.96\pm0.93\%$ \\ \cline{2-6} 
 & Adversarial & $48.37\pm0.99\%$ & $58.92\pm0.97\%$ & $59.45\pm1.00\%$ & $56.33\pm0.98\%$ \\ \hline
\multirow{3}{*}{Matching Nets~\cite{vinyals2016matching}} & Clean & $62.95\pm0.46\%$ & $28.14\pm0.37\%$ & $62.58\pm0.49\%$ & $47.14\pm0.45\%$ \\ \cline{2-6} 
 & $40\%$ & $54.39\pm0.48\%$ & $28.64\pm0.36\%$ & $57.86\pm0.48\%$ & $47.01\pm0.48\%$ \\ \cline{2-6} 
 & Adversarial & $29.40\pm0.44\%$ & $32.77\pm0.42\%$ & $53.34\pm0.52\%$ & $46.50\pm0.46\%$ \\ \hline
\multirow{3}{*}{Relation Nets~\cite{sung2018learning}} & Clean & \boldmath{$75.52\pm0.66\%$} & $35.37\pm0.55\%$ & \boldmath{$75.22\pm0.70\%$} & $55.75\pm0.68\%$ \\ \cline{2-6} 
 & $40\%$ & $66.85\pm0.79\%$ & $36.70\pm0.54\%$ & $68.67\pm0.80\%$ & $55.33\pm0.69\%$ \\ \cline{2-6} 
 & Adversarial & $40.46\pm0.88\%$ & $39.82\pm0.57\%$ & $60.52\pm0.82\%$ & $55.50\pm0.69\%$ \\ \hline
\multirow{3}{*}{ADML (Ours)} & Clean & $69.90\pm0.88\%$ & \boldmath{$65.68\pm0.87\%$} & $69.90\pm0.88\%$ & \boldmath{$59.15\pm0.90\%$} \\ \cline{2-6} 
 & $40\%$ & \boldmath{$67.61\pm0.93\%$} & \boldmath{$62.83\pm0.88\%$} & \boldmath{$69.20\pm0.88\%$} & \boldmath{$60.44\pm0.93\%$} \\ \cline{2-6} 
 & Adversarial & \boldmath{$65.26\pm0.98\%$} & \boldmath{$64.18\pm0.86\%$} & \boldmath{$61.93\pm0.95\%$} & \boldmath{$59.80\pm0.84\%$} \\ \hline
\end{tabular}
}
\end{table*}

\end{document}